\documentclass[letterpaper, 10 pt, conference]{ieeeconf} 
\IEEEoverridecommandlockouts                             
\usepackage{color}
\usepackage{graphicx}
\usepackage{mathtools} 
\usepackage{siunitx}
\usepackage{adjustbox}
\usepackage{array}

\newcolumntype{R}[2]{%
    >{\adjustbox{angle=#1,lap=\width-(#2)}\bgroup}%
    l%
    <{\egroup}%
}

\newcommand\reviewed[1]{\textcolor{black}{#1}}

\title{\LARGE \bf
Fabric Pneumatic Artificial Muscle-Based Head-Neck Exosuit: \\ Design, Modeling, and Evaluation
}
\author{Katalin Schäffer$^{1,2}$, Ian Bales$^{3}$, Haohan Zhang$^{3}$, and Margaret McGuinness$^{1}$
\thanks{This work was supported, in part, by the National Science Foundation (CBET 2240508).}
\thanks{$^{1}$Department of Aerospace and Mechanical Engineering, University of Notre Dame, Notre Dame, IN 46556, USA. {\tt\small \{kschaff2, mmcguinness\}@nd.edu}}
\thanks{$^{2}$Faculty of Information Technology and Bionics, Pázmány Péter Catholic University, 1083 Budapest, Hungary.}
\thanks{$^{3}$Department of Mechanical Engineering and Robotics Center, University of Utah, Salt Lake City, UT 84112 USA. {\tt\small haohan.zhang@utah.edu}}
}
\begin{document}

\maketitle
\thispagestyle{empty}
\pagestyle{empty}

\begin{abstract}
\textcolor{black}{
Wearable exosuits assist human movement in tasks ranging from rehabilitation to daily activities; specifically, head-neck support is necessary for patients with certain neurological disorders. Rigid-link exoskeletons have shown to enable head-neck mobility compared to static braces, but their bulkiness and restrictive structure inspire designs using ``soft" actuation methods.
In this paper, we propose a fabric pneumatic artificial muscle-based exosuit design for head-neck support. We describe the design of our prototype and physics-based model, enabling us to derive actuator pressures required to compensate for gravitational load. Our modeled range of motion and workspace analysis indicate that the limited actuator lengths impose slight limitations (83\% workspace coverage), and gravity compensation imposes a more significant limitation (43\% workspace coverage). We introduce compression force along the neck as a novel, potentially comfort-related metric. We further apply our model to compare the torque output of various actuator placement configurations, allowing us to select a design with stability in lateral deviation and high axial rotation torques. The model correctly predicts trends in measured data where wrapping the actuators around the neck is not a significant factor. Our test dummy and human user demonstration confirm that the exosuit can provide functional head support and trajectory tracking, underscoring the potential of artificial muscle–based soft actuation for head–neck mobility assistance.
}
\end{abstract}

\section{Introduction} \label{sec:Introduction}

The development of wearable robots for human movement assistance is a fast-growing research area. Applications span from rehabilitation to motor function augmentation and support in activities of daily living (ADLs). In addition to assisting the upper and lower extremities, there is a need for devices that can provide head–neck support, particularly for dropped head syndrome resulting from various neurological disorders, such as amyotrophic lateral sclerosis~\cite{burke2025cervical}.

Rigid-link neck exoskeletons have demonstrated their ability to restore head-neck mobility and prove to be a more functional solution than static orthoses~\cite{demaree2024preliminary}. However, their substantial bulk, weight, and restrictive kinematic structures hinder everyday use~\cite{bales2024kinematic}. These limitations have motivated the development of alternative approaches based on soft actuation principles, such as cable-driven mechanisms, which offer lighter, more compact, and more compliant assistance, with greater head-neck range of motion necessary for ADLs~\cite{bales2024kinematic}.

\begin{figure}[tb]
    \centering
    \includegraphics[width=\columnwidth]{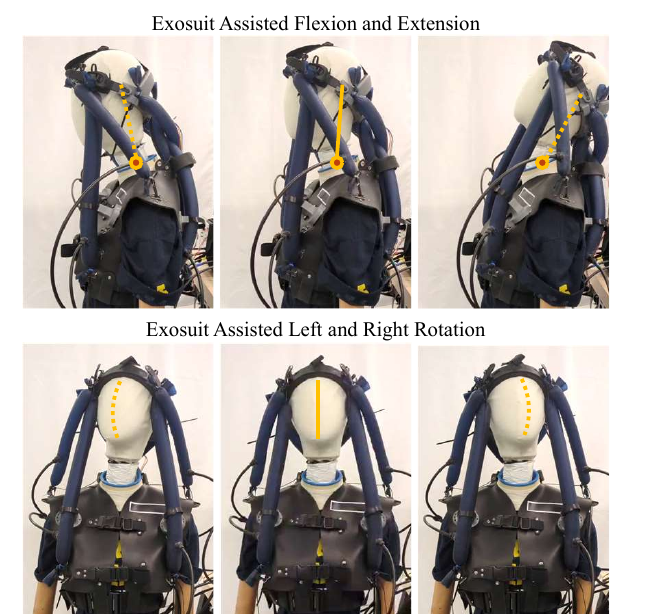}
    \vspace{-0.7cm}
    \caption{Demonstration of our fabric pneumatic artificial muscle-actuated head-neck exosuit assisting flexion-extension movement and axial rotation movement. The exosuit is able to support the anatomically accurate head weight of the test dummy while moving the head along these two degrees of freedom. The exosuit can also assist with lateral deviation.}
    \label{fig:GlamorShot}
    \vspace{-0.6 cm}
\end{figure}

In a cable-driven soft robotic suit design, cables are placed across the biological joints (e.g., cervical spine) to generate moments that rotate the joints when being pulled by actuators (e.g., electric motors). Combining the use of Bowden tubes, the actuation can be placed remotely from the biological joints, thus increasing the portability of cable-driven designs~\cite{xiloyannis2021soft}. However, the cables cannot be placed too close to the body to avoid shearing the skin. Collision with the body will also bend the cables, which undesirably changes the robot kinematic model due to the fundamental assumption of each cable force being directed along the vector formed by the attachment points~\cite{bales2024kinematic}. Additionally, cables generally cannot be placed to cross each other to avoid interference, despite the fact that such configurations increase the ability to generate higher moments about the biological joints. Furthermore, Bowden cable transmission suffers from low mechanical efficiency due to friction loss and compliance of the Bowden sheaths, thereby limiting the accuracy and stability~\cite{xiloyannis2021soft}.

\begin{table*}[tb]
\caption{Summary of Wearable Robotic Head-Neck Device Designs\label{tab:exo_comparison}}
\vspace{-0.3 cm}
\centering
\begin{tabular}{|>{\centering\arraybackslash}p{3.0cm}|>%
{\centering\arraybackslash}p{1.4cm}|>{\centering\arraybackslash}p{9.2cm}|>{\centering\arraybackslash}p{1.8cm}|}
\hline
\textbf{Name and Actuation} & \textbf{Active DoF}& \textbf{Structure and Compliance} & \textbf{Weight} \\
\hline
\hline
fPAM-actuated exosuit (our design) & 3 & Seven fPAM actuators and separate base; design includes routing points and underactuation & 1.2 kg \newline (4.5 kg base)\\
\hline
Cable-driven exosuit~\cite{bales2024kinematic} & 3 & Five cables connect between a shoulder vest and a head piece to apply a moment on the head; cables are remotely actuated using electric motors; compliance is from force control & 1.5 kg\\
\hline
Rigid-link exoskeleton~\cite{demaree2023structurally} & 3& Three-legged revolute-revolute-spherical parallel rigid linkage mechanism optimized to mimic 3D human head-neck movements; actuated by three servomotors mounted on the shoulders; no built-in compliance & 1.2 kg\\
\hline
Elastic neck exoskeleton~\cite{torrendell2024neck} & 2 & Compliant rod mechanism supports the weight of the head; the physical compliance of the rod can be adjusted via motors; has a passive degree of freedom for axial rotation & 2.5 kg  \\
\hline
\end{tabular}\\
\vspace{-0.6 cm}
\end{table*}

In contrast, pneumatic artificial muscles offer a much higher power-to-weight ratio~\cite{thalman2020review}, experience less friction, and allow more flexible routing due to their intrinsic compliance. Specifically, when made of fabric, the muscles can be placed close to the body, and they can move through routing points. 
At the same time, pneumatic artificial muscles present challenges, such as limited contraction, restricted sensing, and slower pneumatic response.
\reviewed{Navigating these trade-offs, pneumatic actuators have been successfully applied in exosuit designs for both upper and lower limbs~\cite{Chen2025glove,Ferroni2025wrist,Pulvirenti2025knee,Kamimura2026hip}. Our goal is to address actuator-specific limitations while exploring the unique capabilities of pneumatic muscles in head–neck assistance.}

In this paper, we introduce an exosuit for the head-neck (Fig.~\ref{fig:GlamorShot}) using soft, fabric pneumatic artificial muscles (fPAMs)~\cite{NaclerioRAL2020}. To the best of our knowledge, this is the first exosuit using pneumatic artificial muscles developed for head-neck mobility, thereby providing the missing knowledge as to how such novel artificial muscles can be leveraged to promote the mobility of the head-neck. Table~\ref{tab:exo_comparison} includes a comparison of our fPAM-based design to other wearable robotic devices designed to assist with head-neck motions, sampled from various actuation categories. 

\textcolor{black}{
In the following sections, we first describe the design of our prototype, followed by modeling of actuator pressures to compensate for the gravitational moment acting on the head. We assess the exosuit's feasibility to reach configurations based on fPAM length, gravity compensation, and limited compression force along the neck. Based on these metrics, we evaluate a visual target workspace and range of motion (RoM). We evaluate various actuator placement strategies via both modeling and measured results. Finally, we demonstrate that the exosuit is capable of moving the head of a test dummy and human user through trajectory tracking.
}

\vspace{-0.2cm}
\section{Design} \label{sec:Design}
The intended purpose of the neck exosuit is to assist with head-neck movements for individuals with head-neck mobility limitations due to reduced neck actuator strengths.
The main consideration is to be able to support the head against gravity at various orientations throughout the range of motion of the head-neck. To this end, a physical design was realized using the fPAM actuation method. In this section, we introduce the overall structure of the exosuit prototype, followed by specification of the artificial muscle placement.

\vspace{-0.1cm}
\subsection{Exosuit Structure}
\begin{figure*}[tb]
    \centering
    \includegraphics[width=\textwidth]{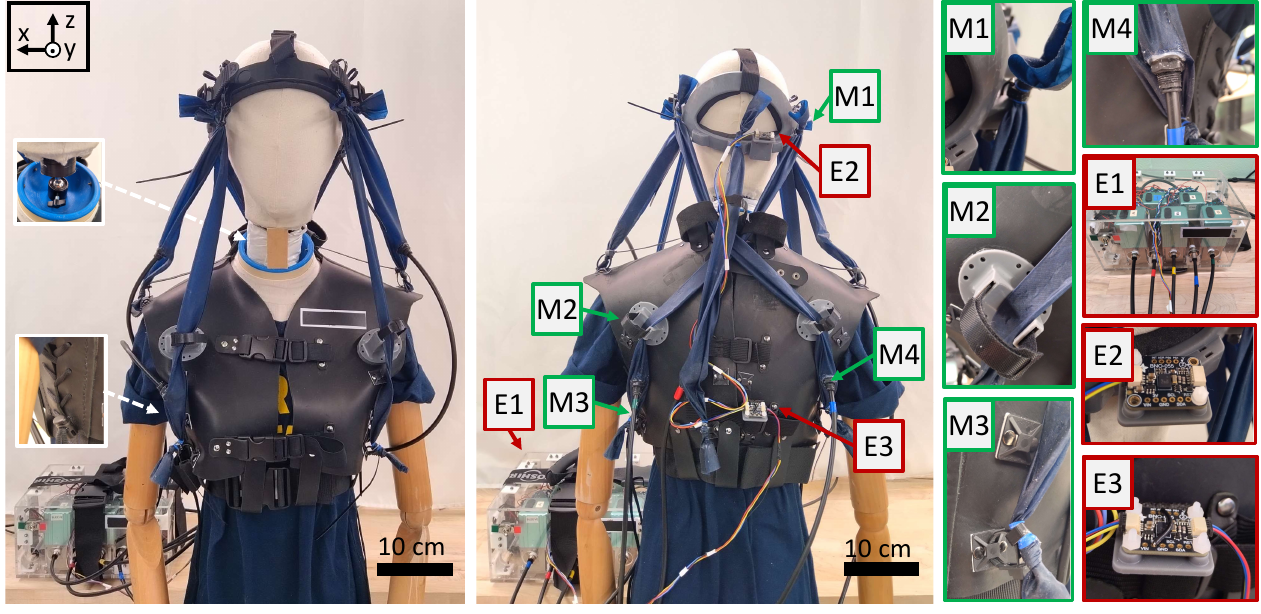}
    \vspace{-0.8cm}
    \caption{Front and back view of the exosuit prototype, with key electrical and mechanical components highlighted. Insets to the front view show the neck joint and side tightening structure of the vest, structures that are hidden in the main pictures. (E1-3) Close-up pictures of the off-board base and the IMU sensors, the electrical components. (M1-3) Actuator head mounting, routing, and torso mounting points, respectively. (M4) Tube (originating from the base) connecting to the actuator.}
    \label{fig:Design}
    \vspace{-0.6 cm}
\end{figure*}

As illustrated in Fig.~\ref{fig:Design}, the exosuit consists of a head piece, a leather vest, actuator mounts, seven fPAM actuators, two inertial measurement units (IMU), and an off-board exosuit base \reviewed{with} pressure regulators and circuitry components.

To demonstrate and conduct bench testing of this prototype, a mannequin with a spherical neck joint (Universal 1/4" Swivel Mini Ball Head Screw Tripod Mount, EszkozTA) was built, as shown in Fig.~\ref{fig:Design}. Additionally, the weight of the mannequin's head was adjusted to 4.4~kg to match the weight of the biological head~\cite{yoganandan2009physical} by filling sand into the cavity. 

The exosuit is attached on the user at the shoulders and head.
The structures to attach the device to the body are similar to the ones used 
in~\cite{bales2024kinematic}. The 3D-printed head piece (Grey Resin, Formlabs and Black Tough PLA, Ultimaker) is mounted with foam padding. A ratcheting mechanism (Toe Sawblade and Ratchet, Union Binding Company) adjusts the head piece on the head. The head piece provides mounting for the fPAM endpoints on the head.

The torso attachment is a customized vest made of leather. Front straps and buckles, as well as side laces, are used to close the vest and secure it to the torso. On the vest, there are arc-shaped, 3D-printed routing structures reinforced with hook-and-loop tapes (M2 in Fig.~\ref{fig:Design}) and plastic zip tie mounts (Zip Tie Adhesive Mounts, Rythcraft) fastened by rivets (M3 in Fig.~\ref{fig:Design}) for anchoring the fPAMs on the vest. 

The fPAMs were fabricated from silicone-coated ripstop nylon fabric (1.1 oz MTN Silnylon 6.6, Ripstop by the Roll) and sealed by glue (Sil-Poxy, Smooth-On). Their knotted ends were mounted using 3D-printed rings and adjustable zip ties (M1 in Fig.~\ref{fig:Design}), which helps adjust their initial stretch. 
All actuators in the exosuit were fabricated similarly, except their length was adjusted to their placement.

The pneumatic system is operated through the off-board base (E1 in Fig.~\ref{fig:Design}), similar to the base described in~\cite{schaffer2024soft}. Five pressure regulators (QB3, Proportion-Air) supply the set pressures to the fPAM actuators as each of them is connected to one of the fPAM actuators through a push-to-connect fitting (Push-to-Connect Tube Fitting, McMaster-Carr) and 1/4~inch tubes (M4 in Fig.~\ref{fig:Design}). The head orientation relative to the torso is measured by two IMU sensors (BNO055, Adafruit), one on the head (E2 in Fig.~\ref{fig:Design}) and one on the vest (E3 in Fig.~\ref{fig:Design}). The IMUs and regulators are synchronized via a microcontroller (Arduino Mega, Arduino) in the base.

\subsection{Actuator Placement}
The function of each fPAM actuator is as follows:
\begin{itemize}
    \item The two pairs of actuators at the front enable flexion (all inflated together) and left or right lateral deviation (each pair inflated separately) of the head.
    \item The actuator at the back middle enables extension of the head.
    \item The two crossed actuators at the back support extension (inflated together) and left or right axial rotation (each separately) of the head.
\end{itemize}

The actuator placements were chosen to maximize their lengths to alleviate effects of limited contraction of the actuators without confining the range of motion (RoM) when deflected. 
Routing points were added to increase moment arms of the actuators and act as pretension to prevent the actuator attachment points from changing. This placement takes advantage of that the actuators can be crossed or even in contact with each other. Notably, the actuators responsible for the axial rotation are crossed. To improve the lateral stability of the head, two short actuators were added at the front and anchored to the shoulders. The benefits of these placement choices are examined through our design evaluation presented in Section~\ref{sec:Design Evaluation}.

The four flexion actuators at the front are underactuated: the two fPAMs at each side are operated from the same pressure regulator. The two actuators in each pair have the same function, but the longer ones 
contribute more to head flexion, and the shorter ones close to the shoulders mostly support the lateral stability.

\section{Actuator Characterization} \label{sec:Actuator Characterization}

To characterize the relationship between the input pressure and the output force of the actuator, we conducted a tensile testing experiment on one of our fabricated fPAMs.

Using the method described in~\cite{NaclerioRAL2020,schaffer2024soft}, we calibrated the parameters for our fPAM actuators, which used fabric from a different manufacturer. 
We also improved the previous method in tensile testing by sampling through more pressure levels and adding a mount for the fPAM to secure the endpoints, resulting in more accurate length measurement.

\begin{figure}[tb]
    \centering
    \includegraphics[width=\columnwidth]{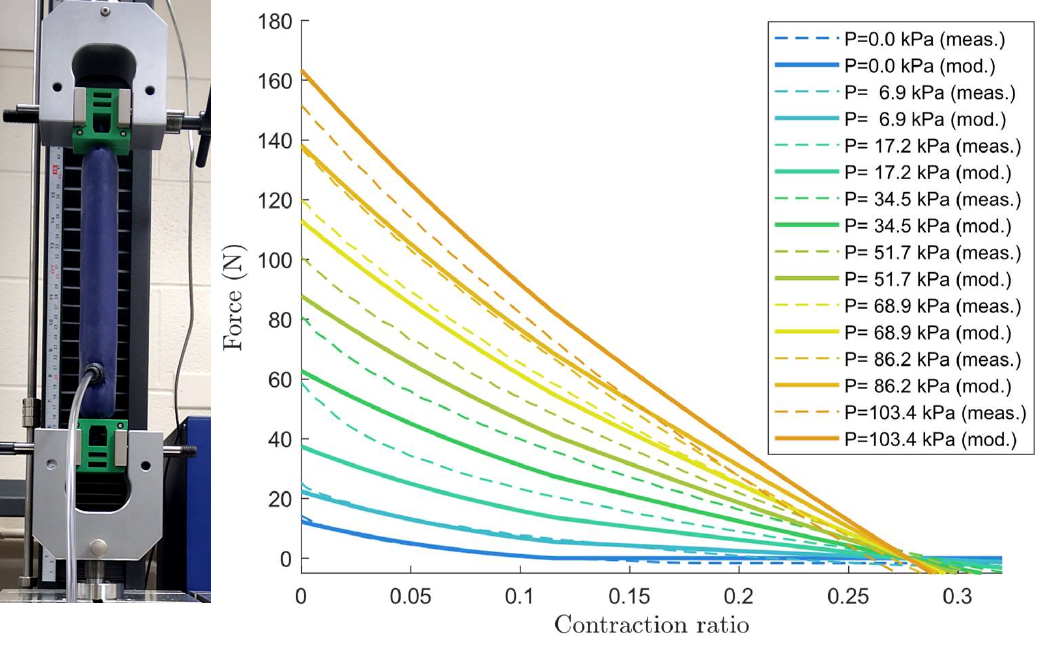}
    \vspace{-1.0cm}
    \caption{Tensile testing setup for characterizing the pressure input and force output relationship of the actuator. The test setup with the fPAM endpoint mounts is shown on the left. The plot on the right shows the measured force data with dashed lines and the modeled force (using Eqn.~\ref{eq:force}) with continuous lines. The colors correspond to various pressure levels ranging from 0~kPa to 103.4~kPa.}
    \label{fig:fPAMForce}
    \vspace{-0.7
    cm}
\end{figure}

Specifically, the endpoints were secured in a tensile testing machine through 3D-printed mounts, and the actuator was stretched repeatedly while the input pressure was kept constant. The measurement was cycled through a set of pressure levels (Fig.~\ref{fig:fPAMForce}) while an inline load cell (SM-1000-961, ADMET) was used to record the force data. This data was used to fit the parameters of the force equation introduced in Section~\ref{subsec:Static Modeling} (Eqn.~\ref{eq:force}).

\section{Exosuit Modeling} \label{sec:Exosuit Modeling}
In this section, we present a model
to compute the actuator pressure to support given head poses. The model assumes  a quasi-static system. 

\subsection{Static Modeling} \label{subsec:Static Modeling}

The user's head-neck is modeled as an inverted pendulum that allows for 3D rotations. The head-neck is assumed to be a moving link, weighing 4.6~kg, with the center of mass (CoM) located 17~cm from the spherical joint that connects to the fixed torso link (based on average human head parameters from the literature~\cite{yoganandan2009physical}). The assigned body-fixed reference frame has its origin at the center of rotation. Similarly to its assigned orientation in~\cite{Kim2010Gaze2HeadAngles}, the \textit{x} axis points towards the right shoulder, the \textit{y} axis points forward along the horizontal plane in the resting position, and the \textit{z} axis points upward along the neck, forming a right-handed coordinate frame (as shown in Fig.~\ref{fig:Design}). Gravity is assumed to apply in the vertical direction in the global inertial frame.

\textcolor{black}{
The head orientation is described using Euler angles corresponding to a body-fixed \textit{x-y-z} sequence. We refer to head flexion-extension (FE) as rotation about the body \textit{x}-axis, lateral deviation (LD) as rotation about the body \textit{y}-axis, and axial rotation (AR) as rotation about the body \textit{z}-axis.
}

The placement of each fPAM actuator is described with the location of the mounting point on the head, as well as the routing and the mounting point on the vest. These points were manually measured on the mannequin. Our model assumes that each actuator runs between these points in straight lines. The force ($F$) that an fPAM applies at a given level of contraction ($\epsilon$) can be calculated based on Eqn.~\ref{eq:force}, which was derived using a method introduced in~\cite{NaclerioRAL2020}.

\begin{equation}
\label{eq:force}
\begin{aligned}
F=&\pi P(\frac{1}{\sin(\alpha_0)^2}-\frac{3(\epsilon-1)^2}{\tan(\alpha_0)^2})r_0^2 + &\} F_{ideal}(\epsilon)P \\
&+ p_3\epsilon^3+p_2\epsilon^2+p_1\epsilon+p_0, &\}F_{elastic}(\epsilon)\\
\text{where }\\
\epsilon =&(L_0 -L)/L_0 \,\, .\\
\end{aligned}
\end{equation}

In Eqn.~\ref{eq:force}, $P$ is the actuator pressure, $L$ represents the configuration-dependent fPAM length, and $L_0$ is the fully stretched length of the actuator, which was manually measured for each fPAM in the exosuit. The other parameters (Table~\ref{tab:force_parameters}), such as the initial radius ($r_0$), the initial weave angle ($\alpha_0$) of the fPAM, and the coefficients of the polynomial approximation of the elastic force ($p_i, i=0,...,3$) were obtained from tensile testing on one fPAM actuator used in this exosuit, as described in Section~\ref{sec:Actuator Characterization}.

Each actuator applies a tensile force ($\mathbf{{f}_{i}}$) along the lines formed by mounting points between the head and torso with magnitude $F_i$ given by Eqn.~\ref{eq:force}. These forces collectively apply a moment on the head about the center of rotation of the neck joint.
This moment and the tensile forces applied by the fPAM actuators can be formulated using Eqn.~\ref{eq:fPAM_torque}: 

\begin{equation}
\label{eq:fPAM_torque}
    \mathbf{\tau_{fPAM}} = \mathbf{J} \mathbf{f_{fPAM}}^T \,\,,\\
\end{equation}
where $\mathbf{f_{fPAM}} = 
    \begin{bmatrix}
        F_{1} & \cdots & F_{7}
    \end{bmatrix}$
is a vector of tension magnitudes of all seven actuators. Matrix $\mathbf{J}$ is the force Jacobian which takes the form of $ \mathbf{J}= 
    \begin{bmatrix}
        \mathbf{b}_{1} \times \hat{\mathbf{c}}_{1} & \cdots & \mathbf{b}_{7} \times \hat{\mathbf{c}}_{7}
    \end{bmatrix}$
where $\mathbf{b}_{i}$ and $\hat{\mathbf{c}}_{i}$ represent the position vector of each actuator attachment on the head piece and the direction vector of each tensile force, respectively.

\textcolor{black}{
To support the weight of the head at an orientation, the required moment applied by the fPAM actuators must balance the gravitational moment at that orientation. With the inverted pendulum model, we are able to estimate this gravitational moment ($\mathbf{\tau_{grav.}}$) with known model parameters. In addition to applying a moment on the head, the tensile forces in the fPAM actuators result in a compression force along the neck. If this compression force is too high, it can result in discomfort of the user. To quantify this force, we define the compression force along the neck ($F_c$), which can be computed as the sum of dot products of the fPAM forces with the position vector of the CoM of the head.
}

\begin{table}[tb]
\caption{\reviewed{fPAM force parameters derived from tensile testing}\label{tab:force_parameters}}
\vspace{-0.3 cm}
\centering
\begin{tabular}{||p{0.9 cm} p{0.9 cm} p{0.9 cm} p{0.9 cm} p{0.9 cm} p{0.9 cm}||}  
\hline
$r_0$ [m] & $\alpha_0$ & $p_0$ & $p_1$  & $p_2$ & $p_3$\\
\hline
0.0136 & 37.0 & 12.3 & -182.9  &  791.3&  -1121.4\\
\hline
\end{tabular}\\
\vspace{-0.5 cm}
\end{table}

\subsection{Pressure Estimation for Gravity Compensation} \label{subsec:GravityCompensation}

To find the required actuator pressures where the gravitational and fPAM torques are equal, we reformulate Eqn.~\ref{eq:fPAM_torque} to be a linear expression of pressures instead of forces. Since the first four actuators form two pairs of actuators, each receiving one pressure input, the formulation changes as follows for a coefficient matrix $A$:

\vspace{-2mm}
\begin{equation}
\label{eq:coefficient_matrix}
\begin{aligned}
    &\mathbf{\tau_{fPAM}} -\mathbf{\tau_{elastic}} = A \mathbf{p}^T\\
    &\text{where }\\
    &A = 
    [\sum_{i=1}^{2} \mathbf{b}_{i} \times F_{ideal,i}\hat{\mathbf{c}}_{i} \,
        \sum_{i=3}^{4} \mathbf{b}_{i} \times F_{ideal,i}\hat{\mathbf{c}}_{i} \\
        &\quad  \quad  \mathbf{b}_{5} \times F_{ideal,5}\hat{\mathbf{c}}_{5} \:
        \cdots \:
        \mathbf{b}_{7} \times F_{ideal,7}\hat{\mathbf{c}}_{7}],\\
    &\mathbf{p} = 
    \begin{bmatrix}
        P_{1} & \cdots & P_{5}
    \end{bmatrix}\\
\end{aligned}
\end{equation}

In Eqn.~\ref{eq:coefficient_matrix}, $\mathbf{\tau_{elastic}}$ denotes the torques applied by the elastic fPAM force calculated from Eqn.~\ref{eq:fPAM_torque} when $\mathbf{p}=0$.

We used the MATLAB built-in function,  \textit{lsqnonneg()} to compute the pressures for gravity compensation where the desired torques are $\mathbf{\tau_{des}}= -\mathbf{\tau_{elastic}}-\mathbf{\tau_{grav}}$. We included a pressure minimization term with the coefficient $\omega=$0.06 by changing the coefficient vector and the desired torque input as in Eqn.~\ref{eq:new_coeffs}.

\vspace{-2mm}
\begin{equation}
\label{eq:new_coeffs}
    \tilde{A} = 
    \begin{bmatrix}
        A \\[6pt]
        \omega \, I_{5 \times 5}
    \end{bmatrix},
    \qquad
    {\mathbf{\tilde{\tau}}_{des}} = 
    \begin{bmatrix}
        \mathbf{\tau}_{des} \\[6pt]
        \mathbf{0}_{5 \times 1}
    \end{bmatrix}.
\end{equation}

\vspace{2mm}
\section{Model-Based Workspace Evaluation} \label{sec:Model-Based RoM and Workspace Evaluation}

\textcolor{black}{
In this section, we define a visual-target workspace for head-neck orientations and evaluate the workspace and range of motion (RoM) given feasible actuator length, gravity compensation, and compression force limits in the neck.
}

\subsection{Feasibility Conditions} \label{subsec:Feasibility Conditions}

Using the model described in Section~\ref{sec:Exosuit Modeling}, we evaluate the exosuit's capability to support various static head configurations. We define three conditions that the exosuit should meet in a given configuration:
\textcolor{black}{
\begin{itemize}
    \item Configuration is reachable based on the fPAM lengths;
    \item Gravity compensation is feasible within pressure limits; 
    \item Compression force is limited to ensure comfort.
\end{itemize}
}

The ``reachable" condition is met if all fPAM actuator lengths are less than their fully stretched length, therefore all contraction ratios are nonnegative. 

The ``gravity compensation" condition is met if feasible solutions exist for the pressures, as discussed in Section~\ref{subsec:GravityCompensation}. The configuration is considered supported if the exosuit torque is within 25\% of the gravitational torque~\cite{bales2024kinematic} or if the error is less than 0.01~N-m. The solver provides positive values for pressures, and we exclude solutions that exceed the 138~kPa limit of the pressure regulators.

\begin{figure}[tb]
    \centering
    \includegraphics[width=\columnwidth]{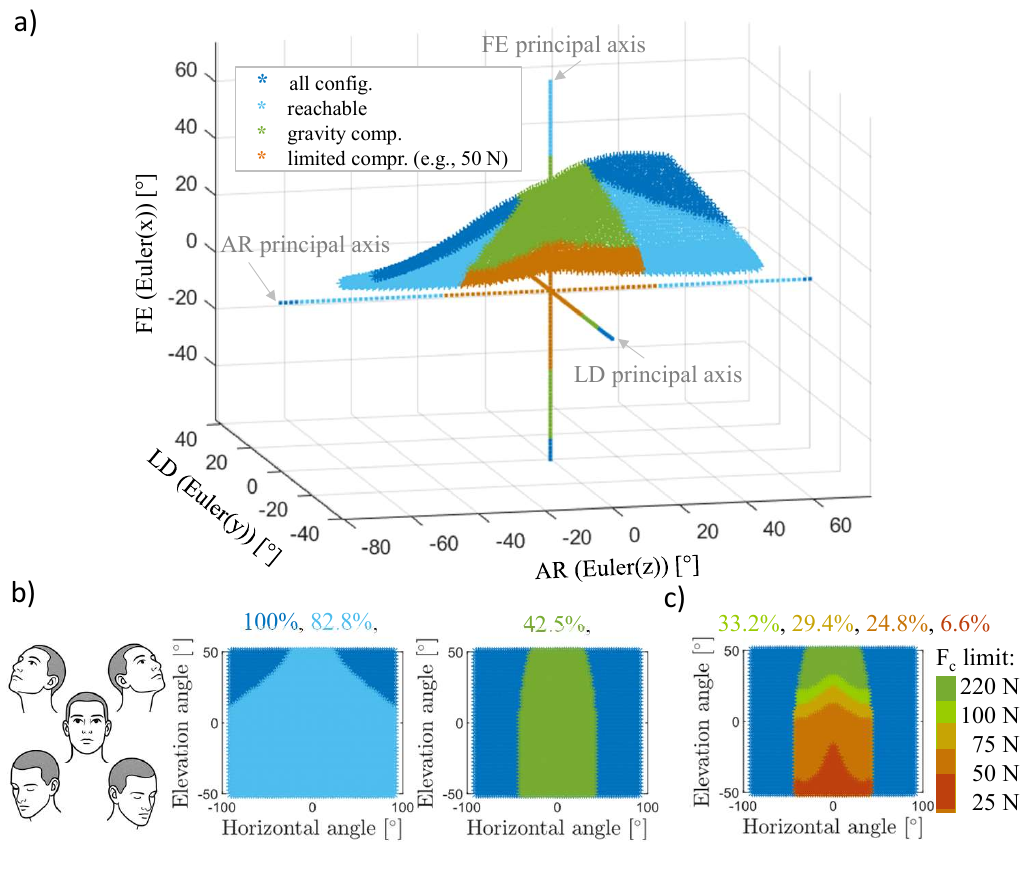}
    \vspace{-1.1cm}
    \caption{
    \textcolor{black}{
    Modeled range of motion and workspace. (a) The principal rotation axes representing range of motion and the visual target workspace configurations in Euler angles (body-fixed \textit{x-y-z}). The colors illustrate whether the exosuit meets the defined evaluation conditions. (b) Illustration of the head orientations corresponding to the 2D visual target workspace, and the workspace plotted under each condition. (c) The workspace changes when enforcing various limits on the compression force ($F_c$).}}
    \label{fig:Workspace}
    \vspace{-0.7 cm}
\end{figure}

Finally, the ``limited compression" condition refers to the case when the computed compression force at the optimized pressures is below a threshold. We examined effects of the levels of thresholds on the resulting workspace for flexion-extension (FE) and axial rotation (AR).

\subsection{Range of Motion and Workspace Evaluation} \label{subsec:Range of Motion and Workspace Evaluation}

We evaluated these three conditions along the primary axes of rotation to explore the exosuit-supported range of motion. We iterated through the full biological RoM as defined in~\cite{Ferrario2002RoM} using 100 data points. Fig.~\ref{fig:Workspace}(a) shows the principle axes for RoM, and Table~\ref{tab:RoM} provides the computed RoM values for each angle from the resting configuration (i.e., with the other two angles set to zero).

To explore how the exosuit performs in head configurations that occur during functional tasks, we define a functional workspace. This workspace is a range of head configurations that corresponds to looking at targets positioned around a person horizontally and vertically at various angles. We set the workspace as $\pm$90$^\circ$ horizontal orientation angles and $\pm$50$^\circ$ vertical elevation angles and iterated through combinations of these angles with 2.5$^\circ$ steps. The conversion from target angles to head orientation in Euler angles was performed following the formulation in~\cite{Kim2010Gaze2HeadAngles}.

Fig.~\ref{fig:Workspace}(a) shows the mapped workspace in Euler angles, while Fig.~\ref{fig:Workspace}(b-c) shows the workspace as a function of target orientations. Here, each condition is illustrated separately, and their workspace coverage is given in percentage.

\textcolor{black}{
The results show that the reachability condition slightly reduces the workspace to 83\% of all configurations tested and most strongly limits the RoM in lateral deviation. Gravity compensation further reduces the workspace to 43\%, with the largest RoM reduction occurring along axial rotation. These findings suggest that future design improvements should prioritize applying higher torques along the latter degrees of freedom. The compressive force analysis provides insight into neck-loading forces that may cause discomfort. Depending on the limit, this can lead to either a small or substantial workspace reduction. Future work should determine whether a specific limit must be defined within the design.
}

\begin{table}[tb]
\caption{Modeled Range of Motion From Resting Configuration \label{tab:RoM}}
\vspace{-0.3 cm}
\centering
\begin{tabular}{||p{1.8 cm}| p{1.6 cm}| p{1.6 cm}| p{1.6 cm}||}  
\hline
     & FE RoM [$^\circ$] (and \%)  & LD RoM [$^\circ$] (and \%) & AR RoM [$^\circ$] (and \%)\\
\hline
Biological & -59.5,+73.7 (100\%) & -40.9,+43.1 (100\%) & -80.8/77.7 (100\%) \\
\hline
Reachable & -48.7,+73.7 (92\%) & -29.9,+30.4 (72\%) & -74.4,74.5  (94\%)\\
\hline
Gravity comp. & -51.4,+46.8 (74\%) & -30.6,+31.2 (75\%) & -31.2,31.3  (40\%)\\
\hline
\end{tabular}\\
\vspace{-0.5 cm}
\end{table}

\section{Design Evaluation} \label{sec:Design Evaluation}

\textcolor{black}{
In this section, we examine actuator placement strategies that differ from the cable-driven structure presented in~\cite{bales2024kinematic} for both front and back actuators. Using the model from Section~\ref{sec:Exosuit Modeling}, we evaluate torques applied by the actuators in their respective degrees of freedom, which helps us choose the preferred design for our application. We also describe the procedure and results for two measurements conducted on the test dummy to validate the modeling results.
}

\subsection{Front Actuator Configurations}

The primary function of the front actuators is to contribute to flexion and lateral deviation. We compare two configurations for this role: one with two actuators (Fig.~\ref{fig:DesignEvaluation}, configuration 1) and one with a single actuator (configuration 2). Since actuator placement is symmetric, we show only the results from the actuators placed on the right side of the user. 

\subsubsection{Evaluation in Modeling}

As shown in Fig.~\ref{fig:DesignEvaluation}(a-b), we evaluate the applied torques for the various actuator placements along flexion-extension (FE) and lateral deviation (LD). In each  case, we assume that the rotational angles about the other two principal axes are zero. In FE, we compute the torque that facilitates flexion (torques with negative sign) and, for LD, we compute the torque in the rightward direction (with positive sign) that is along the actuator's pulling direction. If the actuator is not long enough to let the head reach the given angle, the fPAM force is modeled as infinite, and the resulting torque is not displayed.

For both placements, we compute the integral of the torques that promote flexion and the integral of the torques that promote right lateral deviation. Also, we compute the corresponding angle ranges (Table~\ref{tab:DesignEvaluation}).

\subsubsection{Benchtop Evaluation}

The experiment aims to compare the effect of the two front actuator configurations (Fig.~\ref{fig:DesignEvaluation}, configurations 1 and 2) in lateral deviation. As shown in Fig.~\ref{fig:DesignExperiment}(a), we attached a force sensor to the side of the head, perpendicularly to the sagittal plane, and measured the peak tensile force resulting from the inflation of the right front actuators when pressurized to the maximum pressure of 137~kPa. The head was set to resting orientation, where the dummy head is straight up, facing forward. The torso was fixed using a wooden mount, and the head was supported by a circular support element that prevented it leaning forward but allowed lateral deviation. We repeated the measurement four times each for the one- and two-actuator configurations and included the averaged results in Table~\ref{tab:DesignEvaluation}.

\subsubsection{Results}

Both the modeling results and the experimental evaluation show that adding the short front actuators (configuration 1 versus configuration 2) helps to apply higher torques in LD. The model indicates that the additional actuator reduces the RoM, but it does not change the flexion torque significantly. For our implemented exosuit, we chose the two-actuator placement (configuration 1) to better stabilize the dummy head laterally during movement.

\subsection{Back Actuator Configurations}

The primary function of the two crossed back actuators in the exosuit prototype is to promote axial rotation of the head due to their diagonal placement. We hypothesize that crossing results in greater moment arm than designs where these two actuators are not allowed to cross and therefore they are less tilted compared to vertical direction. For example, in a previous exosuit design using cables~\cite{bales2024kinematic}, the authors used a similar cable-actuator arrangement on the back of the head. 
However, because cables are not allowed to cross, the previous cable design had low torque output for the axial rotation motion of the head.

\subsubsection{Evaluation in Modeling}

\begin{figure}[tb]
    \centering
    \includegraphics[width=\columnwidth]{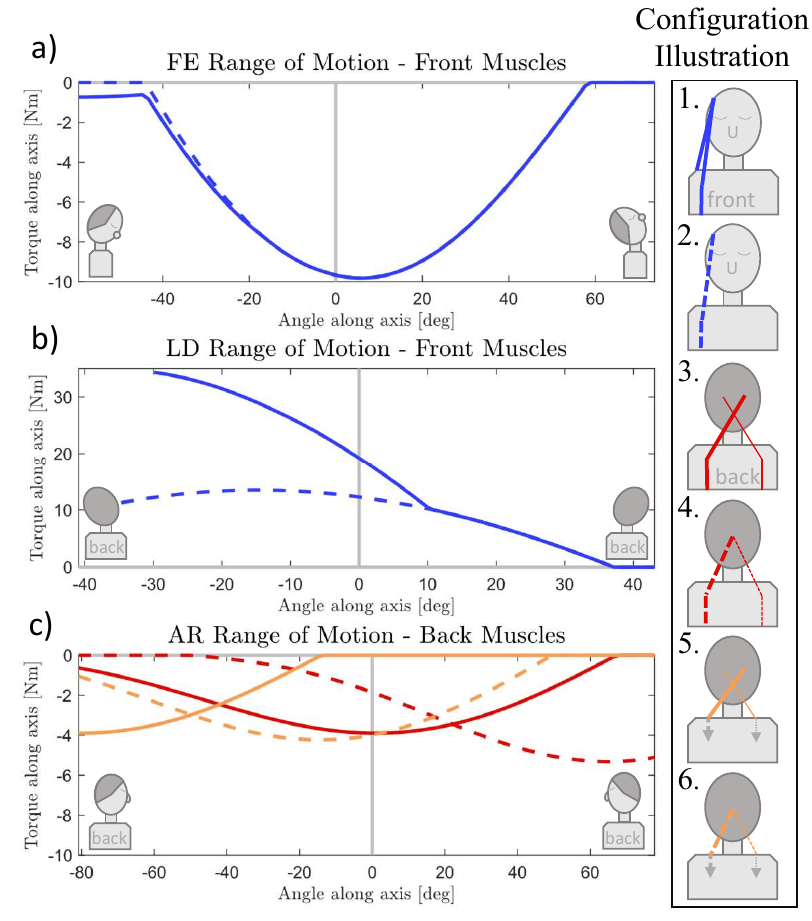}
    \vspace{-0.8cm}
    \caption{Modeled fPAM torques in (a) flexion-extension (FE), (b) lateral deviation (LD), and (c) axial rotation (AR) for various actuator configurations. The configurations are: (1) two front actuators, (2) one front actuator, (3) crossed back actuators anchored to the back of the vest, (4) upside-down ``V" back actuators anchored to the back of the vest, and (5 and 6) the same placements as (3 and 4) but anchored to the front of the vest.}
    \label{fig:DesignEvaluation}
    \vspace{-0.6 cm}
\end{figure}

We evaluate two possible variations of the back actuators' placement. First, the actuators can be placed on the headpiece crossed in an ``X" shape, or they can meet at the mid-point, forming an upside down ``V" shape. For the second case, the modeled anchoring point location on the head is changed to the midpoint on the headpiece. Second, the actuators can be anchored either to the front or to the back of the vest. To implement these configurations, the modeled routing point and anchoring point location on the vest are mirrored across the frontal (\textit{x}-\textit{z}) plane. This gives a total of four combinations (Fig.~\ref{fig:DesignEvaluation}, configurations 3-6).

As shown in Fig.~\ref{fig:DesignEvaluation}(c), we evaluate the torque that the actuators apply in axial rotation (AR) for the various actuator placements. We compute the torque with negative sign that facilitates the right turn of the head, which corresponds to the actuator's pulling direction.
For each configuration, we compute the integral of the torques, along with the corresponding angle ranges (Table~\ref{tab:DesignEvaluation}). Configuration 3 is our implemented prototype structure, and configuration 6 follows the cable-driven structure.

\subsubsection{Benchtop Evaluation}

\begin{figure}[tb]
    \centering
    \includegraphics[width=\columnwidth]{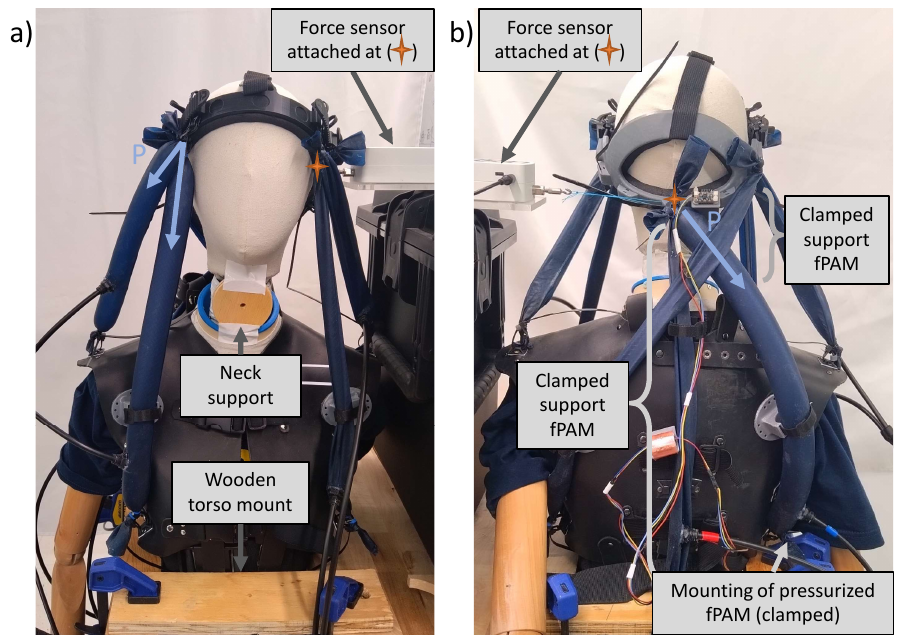}
    \vspace{-0.8cm}
    \caption{Experimental setup for measuring the force applied by (a) the front actuators in lateral deviation (LD), and (b) the back actuators in axial rotation (AR).}
    \label{fig:DesignExperiment}
    \vspace{-0.2 cm}
\end{figure}

This experiment compares the rotational torque for the four back actuator configurations (Fig.~\ref{fig:DesignEvaluation}, configurations 3 through 6). Using the measurement setup in Fig.~\ref{fig:DesignExperiment}(b), we measured the tensile force resulting from the inflation of the actuator (on the right side). The force sensor was attached at the middle, on the back of the headpiece. It was positioned to be perpendicular to the sagittal (\textit{y}-\textit{z}) plane, such that it resisted the axial rotation of the head. Two fPAMs were stretched and clamped as shown in Fig.~\ref{fig:DesignExperiment}(b), to stabilize the head. To ensure the mounting point on the vest was horizontally consistent across all configurations, the end of the fPAM was secured to the wooden torso mount. Then, the applied force was measured for each configuration four times when the actuator was inflated to the maximum pressure. The averaged results of each of the four measurements are in Table~\ref{tab:DesignEvaluation}.

\subsubsection{Results}

Comparing the four placement options of the two actuators at the back, the model evaluation shows that configurations corresponding to our design (configuration 3) and the cable-driven structure in~\cite{bales2024kinematic} (configuration 6) behave similarly. When placed on the right side, both promote the axial rotation of the head in the rightward direction along a large portion of the RoM. The model shows the advantage of the crossed anchoring on the head (configuration 3) over the upside-down ``V" arrangement (configuration 4) when the actuators are anchored on the back.

The measurement results confirm the advantage of the crossed configuration for the back anchored configurations. However, the measurements do not align with the model predictions for the front anchored configurations. We expected that the ratios of the measured forces would be similar to the ratios of the modeled torque magnitudes at 0$^\circ$ (Table~\ref{tab:DesignEvaluation}). The mismatch indicates that the model is not accurate when the fPAM actuators have a more complex routing, in this case having one mounting point at the back and the other in the front. We observed that in this case, the routing of the actuator was bent by the neck, therefore we hypothesize that we need to model the interaction of the actuators with each other and with the neck to provide a reliable prediction.

\begin{table}[tb]
\caption{Model-Based and Experimental Design Evaluation of fPAMs in Various Configurations on the Right Side of the  Head \label{tab:DesignEvaluation}}
\vspace{-0.3 cm}
\centering
\begin{tabular}{||p{1.2 cm} | p{0.42 cm} p{0.4 cm} p{0.4 cm} p{0.4 cm} p{0.4 cm} p{0.4 cm} p{0.4 cm} p{0.4 cm}||}  
\hline
 & 1st (FE) & 2nd (FE) & 1st (LD) & 2nd (LD) & 3rd (AR) & 4th (AR) & 5th (AR) & 6th (AR)\\
\hline
Torque Integr. [Nm] & 630 & 649 & 1147 & 752 & 369 & 360  & 169 & 369\\
\hline
Angle Range [$^{\circ}$] & 116 &  101  & 66 & 75 & 147  & 126 & 96 & 130\\
\hline
$\|\tau\|$ at 0$^\circ$ [Nm] & 9.7 &  9.7  & 19.2 & 12.3 & 3.9  & 1.9 & 0.0 & 4.0\\
\hline
\hline
Measured force [N] & - &  -  & 22.6 & 17.7 & 18.0  & 5.6 & 23.2 & 8.9\\
\hline
\end{tabular}\\
\vspace{-0.5 cm}
\end{table}

\section{Demonstration}
In this section, we demonstrate that the exosuit can move the head along trajectories within the range of motion for both the test dummy and a human user.

\subsection{Controller Architecture}

To achieve angle trajectory tracking along each degree of freedom, we applied an input-output antagonistic feedback similar to the controller applied in~\cite{schaffer2024soft}. The feedback controller derives an error term based on the difference between the desired and current joint angles. Then, along each degree of freedom, the feedback term is added to the pressures of the set of actuators that are defined as ``agonist" and subtracted from the pressures of the ``antagonist" actuators. The sets of agonist/antagonist actuators are pre-defined: The four front actuators are acting against the three actuators in the back in FE, the two crossed back actuators form a pair of antagonistic actuators for AR, and the two sets of front actuators form sets of antagonistic actuators for LD. The controller runs along each degree of freedom to set the desired angle. For the movements in the demonstrations, we did not run controllers that interfered with each other, therefore each actuator received input from only one controller.

\subsection{Trajectory Tracking}

The two commanded trajectories were flexion-extension (FE) and axial rotation (AR) with $\pm$20$^\circ$ from the initial resting position, as these trajectories lie in the gravity compensation workspace. The initial pressures in the actuators were set to 34.5~kPa before starting the tracking. In case of the flexion-extension trajectory, the feedback controller was applied only along the corresponding degree of freedom, while, for the axial rotation trajectory, feedback was applied for flexion-extension and axial rotation to preserve the head's stability. A sinusoidal trajectory with a 25~s period was traced four times for each measurement.

\begin{figure}[tb]
    \centering
    \includegraphics[width=\columnwidth]{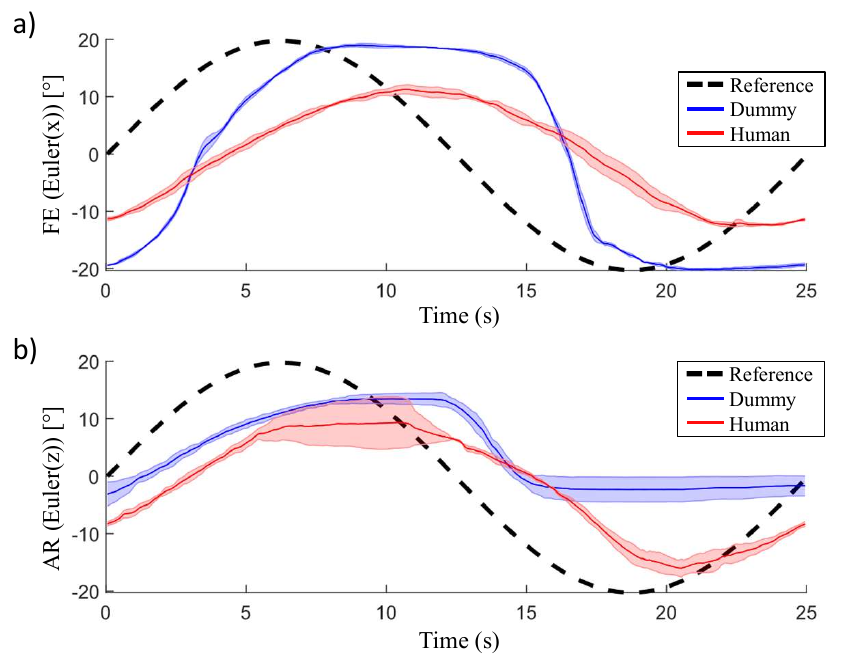}
    \vspace{-0.8cm}
    \caption{
    \textcolor{black}{Results of trajectory tracking with a 20$^\circ$ amplitude and 25~s period sinusoidal reference signal. The tracking was performed along (a) flexion-extension and (b) axial rotation. The experiment was repeated with the exosuit placed on the test dummy (blue) and on a human user (red).}
    }
    \label{fig:TrajectoryTracking}
    \vspace{-0.8 cm}
\end{figure}

The trajectory tracking results for the test dummy and with one healthy participant are shown in Fig.~\ref{fig:TrajectoryTracking}. For each of the trajectories, we computed the time delay using MATLAB's \textit{xcorr} function. We also computed the root mean square error (RMSE) between the measured data and the reference signal when shifted by the computed time delay (Table~\ref{tab:TrajectoryEvaluation}).

During the evaluation, the participant was instructed to relax their neck and let the exosuit move their head throughout the duration of each trajectory. Although we cannot exclude some active movement from the participant, the test confirmed that the exosuit can be applied to a human user. The test dummy is biologically less accurate, especially because the neck has less support, but this demonstration showcased that the exosuit can move the weight of the head completely unsupported.

\begin{table}[tb]
\caption{Trajectory Tracking Evaluation\label{tab:TrajectoryEvaluation}}
\vspace{-0.3 cm}
\centering
\begin{tabular}{||p{1.6 cm} | p{1.2 cm} p{1.2 cm} | p{1.2 cm} p{1.2 cm}||}  
\hline
 & Human (delay) & Human (RMSE) & Dummy (delay) &Dummy (RMSE)\\
\hline
FE trajectory & 4.50 s & 6.60$^{\circ}$ & 4.12 s & 4.60$^{\circ}$\\
\hline
AR trajectory & 2.43 s & 5.96$^{\circ}$ & 2.68 s & 9.24$^{\circ}$\\
\hline
\end{tabular}\\
\vspace{-0.5 cm}
\end{table}





\section{Conclusion} \label{sec:Conclusions} 
We presented the first head–neck exosuit actuated by fabric pneumatic artificial actuators (fPAMs). Compared to a cable-driven neck exosuit design in the literature~\cite{bales2024kinematic}, the fPAM-based design requires maximizing actuator lengths, but allows greater freedom in actuator placement, since the actuators are safe to contact the body and each other.

Through actuator characterization and kinematics, we developed a physics-based model of the exosuit. We derived actuator pressures required to compensate for the gravitational force. We also introduced a visual target workspace and assessed range of motion (RoM) and workspace feasibility considering fPAM length-based reachability, gravity compensation, and compression forces along the neck. The results revealed that gravity compensation is the primary limiting factor, only covering 43\% of the workspace compared to 83\% reachability, with RoM particularly constrained in axial rotation. \reviewed{RoM measured across ADL tasks in~\cite{bible2010RoM_ADL} supports that the range in this direction needs to be extended to fulfill assistive requirements.} We also proposed evaluating compression force along the neck, the practical implications of which for linear contractile actuators require further study. \reviewed{Once safety limits are established, future work should also include the integration of force sensing to ensure that compressive loads remain within safe bounds during operation.} 

The model also enables evaluation of various actuator placement strategies. Experimental results confirmed several model predictions, though discrepancies highlighted unmodeled effects and the need for placement parameter fitting based on experimental measurements. We hypothesize that the most significant unmodeled effect is the actuator–environment interaction, \reviewed{which should be explicitly incorporated and systematically evaluated in future work. We also plan to conduct sensitivity analysis and inter-user anthropometric variability studies.}

Finally, we demonstrated that the proposed exosuit can move the head along flexion–extension and axial rotation using simple antagonistic feedback control, \reviewed{confirming the system’s actuation capability. However, this approach exhibited noticeable delay and tracking errors. Future work will implement more advanced predictive control strategies, such as in~\cite{haggerty2023control}, to facilitate dynamic tracking. We will also conduct user studies in clinical collaboration to thoroughly assess applicability; we will also compare the fPAM, rigid-link, and cable-driven designs using metrics such as positioning error, trajectory smoothness, peak velocity differences, RoM, and mean electromyography amplitudes.}



\section{Acknowledgements} \label{sec:Acknowledgements}
We thank Adballa Ibrahim and Alan Cristian Roque Rivera for their assistance with design and early prototype testing.

\bibliographystyle{IEEEtran}
\bibliography{library_HNE}

@article{xiloyannis2021soft,
  title={Soft robotic suits: State of the art, core technologies, and open challenges},
  author={Xiloyannis, Michele and Alicea, Ryan and Georgarakis, Anna-Maria and Haufe, Florian L and Wolf, Peter and Masia, Lorenzo and Riener, Robert},
  journal={IEEE Transactions on Robotics},
  volume={38},
  number={3},
  pages={1343--1362},
  year={2021},
  publisher={IEEE}
}

@article{torrendell2024neck,
  title={A neck orthosis with multi-directional variable stiffness for persons with dropped head syndrome},
  author={Torrendell, Santiago Price and Kadone, Hideki and Hassan, Modar and Chen, Yang and Miura, Kousei and Suzuki, Kenji},
  journal={IEEE Robotics and Automation Letters},
  volume={9},
  number={7},
  pages={6224--6231},
  year={2024},
  publisher={IEEE}
}

@article{burke2025cervical,
  title={Cervical Collar Satisfaction and Functional Impact in Amyotrophic Lateral Sclerosis: A Survey Study},
  author={Burke, Katherine M and Shea, Cristina and Arulanandam, Vishni and Sullivan, Stacey and Ellrodt, Amy S and MacAdam, Claire and Carney, Kendall and Casagrande, Gabriella and Christiansen, Elizabeth and Paganoni, Sabrina},
  journal={American Journal of Physical Medicine \& Rehabilitation},
  pages={10--1097},
  year={2025},
  publisher={LWW}
}

@article{thalman2020review,
  title={A review of soft wearable robots that provide active assistance: Trends, common actuation methods, fabrication, and applications},
  author={Thalman, Carly and Artemiadis, Panagiotis},
  journal={Wearable Technologies},
  volume={1},
  pages={e3},
  year={2020},
  publisher={Cambridge University Press}
}

@article{NaclerioRAL2020,
  title={Simple, Low-hysteresis, Foldable, Fabric Pneumatic Artificial Muscle},
  author = {N. Naclerio and E. W. Hawkes},
  journal={IEEE Robotics and Automation Letters},
  doi = {10.1109/LRA.2020.2976309},
  year={2020},
  pages={3406-3413},
  volume = {5},
  number = {2}
}

@article{Ferrario2002RoM,
  title={Active range of motion of the head and cervical spine: a three‐dimensional investigation in healthy young adults},
  author = {Ferrario, V. F. and Sforza, C. and Serrao, G. and Grassi, G. and Mossi, E.},
  journal={Journal of Orthopaedic Research},
  year={2002},
  pages={122-129},
  volume = {20},
  number = {2}
}

@article{Kim2010Gaze2HeadAngles,
author = {K. Han Kim and Matthew P. Reed and Bernard J. Martin},
title = {A model of head movement contribution for gaze transitions},
journal = {Ergonomics},
volume = {53},
number = {4},
pages = {447--457},
year = {2010},
publisher = {Taylor \& Francis},
doi = {10.1080/00140130903483713}
}

@ARTICLE{bales2024kinematic,
  author={Bales, Ian and Zhang, Haohan},
  journal={IEEE Robotics and Automation Letters}, 
  title={Kinematic Benefits of a Cable-Driven Exosuit for Head-Neck Mobility}, 
  year={2024},
  volume={9},
  number={12},
  pages={11849-11856},
  doi={10.1109/LRA.2024.3500878}}

@article{demaree2024preliminary,
  title={Preliminary study on effects of neck exoskeleton structural design in patients with amyotrophic lateral sclerosis},
  author={Demaree, David and Brignone, Joseph and Bromberg, Mark and Zhang, Haohan},
  journal={IEEE Transactions on Neural Systems and Rehabilitation Engineering},
  volume={32},
  pages={1841--1850},
  year={2024},
  publisher={IEEE}
}

@inproceedings{demaree2023structurally,
  title={A structurally enhanced neck exoskeleton to assist with head-neck motion},
  author={Demaree, David and Zhang, Haohan},
  booktitle={IEEE International Symposium on Medical Robotics},
  pages={1--7},
  year={2023}
}

@article{yoganandan2009physical,
  title={Physical properties of the human head: mass, center of gravity and moment of inertia},
  author={Yoganandan, Narayan and Pintar, Frank A and Zhang, Jiangyue and Baisden, Jamie L},
  journal={Journal of Biomechanics},
  volume={42},
  number={9},
  pages={1177--1192},
  year={2009},
  publisher={Elsevier}
}

@article{schaffer2024soft,
  title={Soft wrist exosuit actuated by fabric pneumatic artificial muscles},
  author={Sch{\"a}ffer, Katalin and Ozkan-Aydin, Yasemin and Coad, Margaret M},
  journal={IEEE Transactions on Medical Robotics and Bionics},
  volume={6},
  number={2},
  pages={718--732},
  year={2024},
  publisher={IEEE}
}

@article{bible2010RoM_ADL,
  title={Normal functional range of motion of the cervical spine during 15 activities of daily living},
  author={Bible, Jesse E and Biswas, Debdut and Miller, Christopher P and Whang, Peter G and Grauer, Jonathan N},
  journal={Clinical Spine Surgery},
  volume={23},
  number={1},
  pages={15--21},
  year={2010},
  publisher={LWW}
}

@article{haggerty2023control,
  title={Control of soft robots with inertial dynamics},
  author={Haggerty, David A and Banks, Michael J and Kamenar, Ervin and Cao, Alan B and Curtis, Patrick C and Mezi{\'c}, Igor and Hawkes, Elliot W},
  journal={Science robotics},
  volume={8},
  number={81},
  pages={eadd6864},
  year={2023},
  publisher={American Association for the Advancement of Science}
}

@article{Ferroni2025wrist,
  title={A soft pneumatic exosuit to assist pronosupination in individuals with spinal cord injury},
  author={Ferroni, Roberto and D’Avola, Gaetano and Mauceri, Daniele Filippo and Pau, Chiara and Sciarrone, Giorgia and Righi, Gabriele and Carpaneto, Jacopo and Gandolla, Marta and Del Popolo, Giulio and Micera, Silvestro and others},
  journal={Advanced Intelligent Systems},
  volume={7},
  number={12},
  pages={e202500124},
  year={2025},
  publisher={Wiley Online Library}
}

@INPROCEEDINGS{Chen2025glove,
  author={Chen, Rui and Leonardis, Daniele and Frisoli, Antonio and Chiaradia, Domenico},
  booktitle={International Conference On Rehabilitation Robotics}, 
  title={A Powerful Customized Fabric-Based Soft Robotic Glove for Assistance and Rehabilitation}, 
  year={2025},
  volume={},
  number={},
  pages={669-674},
  doi={10.1109/ICORR66766.2025.11063169}}

@inproceedings{Kamimura2026hip,
  title={Human-in-the-Loop Control of a Soft Pneumatic Exosuit for Step Width Guidance via Hip Abduction/Adduction},
  author={Kamimura, Jinnosuke and Miyazaki, Tetsuro and Kawashima, Kenji},
  booktitle={IEEE/SICE International Symposium on System Integration},
  pages={358--363},
  year={2026},
}

@article{Pulvirenti2025knee,
  title={A Resistive Soft Robotic Exosuit for Dynamic Body Loading in Hypogravity},
  author={Pulvirenti, Emanuele and Diteesawat, Richard Suphapol and Pavei, Gaspare and Natalucci, Valentina and Hauser, Helmut and Minetti, Alberto and Rossiter, Jonathan},
  journal={Advanced Science},
  volume={12},
  number={47},
  pages={e06057},
  year={2025},
  publisher={Wiley Online Library}
}

\end{document}